\documentclass{article}

\usepackage{microtype}
\usepackage{graphicx}
\usepackage{subfigure}
\usepackage{booktabs} 
\usepackage{algorithmic}
\usepackage{algorithm}
\usepackage[algo2e]{algorithm2e}
\usepackage{graphicx}
\usepackage{multirow}
\usepackage{makecell}
\usepackage{amsmath}
\usepackage{wrapfig}
\usepackage{lipsum}

\usepackage{hyperref}

\usepackage[accepted]{icml2019}

\icmltitlerunning{Submission and Formatting Instructions for ICML 2019}

\begin{document}

\twocolumn[
\icmltitle{ResNet Can Be Pruned 60$\times$: Introducing Network Purification and Unused Path Removal (P-RM) after Weight Pruning}



\icmlsetsymbol{equal}{*}

\begin{icmlauthorlist}
\icmlauthor{Xiaolong Ma}{equal,neu}
\icmlauthor{Geng Yuan}{equal,neu}
\icmlauthor{Sheng Lin}{neu}
\icmlauthor{Zhengang Li}{neu}
\icmlauthor{Hao Sun}{neu2}
\icmlauthor{Yanzhi Wang}{neu}
\end{icmlauthorlist}

\icmlaffiliation{neu}{Department of Electrical and Computer Engineering, Northeastern University, Boston, USA}
\icmlaffiliation{neu2}{Department of Civil and Environmental Engineering, Northeastern University, Boston, USA}

\icmlcorrespondingauthor{Xiaolong Ma}{ma.xiaol@husky.neu.edu}

\icmlkeywords{Machine Learning, ICML}

\vskip 0.2in
]



\printAffiliationsAndNotice{\icmlEqualContribution} 

\begin{abstract}
The state-of-art DNN structures involve high computation and great demand for memory storage which pose intensive challenge on DNN framework resources. To mitigate the challenges, weight pruning techniques has been studied. However, high accuracy solution for extreme structured pruning that combines different types of structured sparsity still waiting for unraveling due to the extremely reduced weights in DNN networks. In this paper, we propose a DNN framework which combines two different types of structured weight pruning (filter and column prune) by incorporating \textit{alternating direction method of multipliers} (ADMM) algorithm for better prune performance. We are the first to find non-optimality of ADMM process and unused weights in a structured pruned model, and further design an optimization framework which contains the first proposed \textit{Network Purification} and \textit{Unused Path Removal} algorithms which are dedicated to post-processing an structured pruned model after ADMM steps. Some high lights shows we achieve 232$\times$ compression on LeNet-5, 60$\times$ compression on ResNet-18 CIFAR-10 and over 5$\times$ compression on AlexNet. We share our models at anonymous link \url{http://bit.ly/2VJ5ktv}.
\end{abstract}
\vspace{-2.2em}
\section{Introduction}
In order to solve the high demand for computation and storage resources of a DNN application, weight pruning~\cite{han2015learning}~\cite{wen2016learning} are developed to facilitate weight compression and computation acceleration. In this work, a structured pruning technique is utilized to compress the DNN models which reduces weight storage and computation, 
and the structured weight matrix storage also has potential advantages for high-parallelism implementation in hardware by eliminating the required weight indices compared with irregular pruning.~\cite{ding2018structured}~\cite{zhang2018adam}

However, the accuracy loss problem in structured pruning is inevitable. By adopting ADMM~\cite{boyd2011distributed}, the original weight pruning problem is decomposed into two subproblems, one solved using stochastic gradient descent as original DNN training, while the other solved optimally and analytically via Euclidean projection~\cite{zhang2018adam}~\cite{ye2019progressive}. ADMM method achieves one of the state-of-art structured weight pruning results, 40$\times$ weight reduction on LeNet-5~\cite{lecun1998gradient} with MNIST~\cite{lecun2015deep}, 20$\times$ on VGG-16~\cite{simonyan2014very} with CIFAR-10~\cite{krizhevsky2009learning} and 4.7$\times$ on AlexNet~\cite{krizhevsky2012imagenet} with ImageNet~\cite{deng2009imagenet} \emph{without post-processing} optimization. 

During post-processing procedure, we find that after model retraining, some weights become less contributing to the network performance. This phenomenon is caused by the shortcoming that ADMM technique lacks the guarantee on solution feasibility (non-optimality) due to the non-convex nature of objective function (loss function). We propose a novel algorithm to detect and remove the redundant weights which slip away from ADMM pruning. Also, we are the first to discover the unused path in a structured pruned DNN model and design a sophisticate optimization framework to further boost compression rate as well as maintain high network accuracy. The contributions of this paper include:
\vspace{-0.8em}
\begin{itemize}
    \item We adopt ADMM for efficiently optimizing the non-convex problem and successfully utilized this method on structured weight pruning.
    \vspace{-0.6em}
    \item We design a novel \textit{Network Purification} and \textit{Unused Path Removal} (P-RM) algorithm focused on post-processing an ADMM structured pruned model to boost compression rate while maintain accuracy.
\end{itemize}

\vspace{-1.2em}
\section{ADMM model compression}
Consider an $N$-layer DNNs, sets of weights of the $i$-th (CONV or FC) layer are denoted by ${\bf{W}}_{i}$, respectively. And the \textit{loss function} associated with the DNN is denoted by $f \big( \{{\bf{W}}_{i}\}_{i=1}^N \big)$. In this paper, $\{{\bf{W}}_{i}\}_{i=1}^N$ characterize the set of weights from layer $1$ to layer $N$. The overall problem is defined by
\vspace{-0.20em}
\begin{equation}
\small
\label{original}
\begin{aligned}
& \underset{ \{{\bf{W}}_{i}\}}{\text{minimize}}
& & f \big( \{{\bf{W}}_{i}\}_{i=1}^N \big),
\\ & \text{subject to}
& & {\bf{W}}_{i}\in {\bf{\mathcal{P}}}_{i}, \; {\bf{W}}_{i}\in {\bf{\mathcal{Q}}}_{i}, \; i = 1, \ldots, N.
\end{aligned}
\end{equation}
Given the value of $\alpha_{i}$, the constraint set is denoted by  ${{\bf{\mathcal{P}}}_{i}=\{{\bf{W}}_{i}|card(supp({\bf{W}}_{i}))\leq\alpha_{i}\}}$, where ``card" refers to cardinality and ``supp" refers to the support set. Elements in ${\bf{\mathcal{P}}}_{i}$ are the solution of ${\bf{W}}_{i}$ satisfying the number of non-zero elements in ${\bf{W}}_{i}$ is limited by $\alpha_{i}$ for layer $i$. The general constraint can be extended in structured pruning such as filter pruning, channel pruning and column pruning.

The standard ADMM regularized optimization steps are shown as follow,
consider a indicator function is utilized to incorporate $\bf{\mathcal{P}_{i}}$ into objective functions, which is
\vspace{-3mm}
\begin{equation}
    \footnotesize
    g_{i}({\bf{W}}_{i})=
    \begin{cases}
        0 & \text {if } {\bf{W}}_{i}\in {\bf{\mathcal{P}}}_{i} \\ 
        +\infty & \text {otherwise}
    \end{cases}
    \quad i = 1, \ldots, N
    \label{object_func}
    \vspace{-4mm}
\end{equation}

Then original problem (\ref{original}) can be equivalently rewritten as 
\vspace{-2.5mm}
\begin{equation}
\footnotesize
\label{admm_form}
\begin{aligned}
& \underset{ \{{\bf{W}}_{i}\}}{\text{minimize}}
& & f \big( \{{\bf{W}}_{i} \}_{i=1}^N \big)+\sum_{i=1}^{N} g_{i}({\bf{Y}}_{i}),
\\ & \text{subject to}
& & {\bf{W}}_{i} = \textbf{Y}_i, \; i = 1, \ldots, N,
\end{aligned}
\vspace{-1mm}
\end{equation}
Auxiliary variables ${\bf{Y}}_{i}$ and dual variables ${\bf{U}}_{i}$ are imported. ADMM decompose problem (\ref{admm_form}) into simpler subproblems and solve subproblems iteratively until convergence. The augmented Lagrangian formation of problem (\ref{admm_form}) is
\vspace{-3mm}
\begin{equation}
\small
f \big( \{{\bf{W}}_{i} \}_{i=1}^N \big)+\sum_{i=1}^{N} \frac{\rho_{i}}{2}  \| {\bf{W}}_{i}-{\bf{Y}}_{i}+{\bf{U}}_{i} \|_{F}^{2}
\label{equ7}
\vspace{-2.2mm}
\end{equation}
The first term in problem (\ref{equ7}) is the differentiable loss function of the DNN, and the second term is a quadratic regularization term of the ${\bf{W}}_{i}$, which is differentiable and convex, and $\|\cdot\|_F^2$ denotes Frobenius norm. As a result, subproblem (\ref{equ7}) can be solved by stochastic gradient descent algorithm~\cite{kingma2014adam} as the original DNN training.

The standard ADMM algorithm~\cite{boyd2011distributed} steps proceed by repeating, for $k = 0, 1,\ldots$, the following subproblems iterations:
\begin{equation}
\small
    \bf{W}_{i}^{k+1} := \underset{ {\bf{W}}_{i}}{\text{arg min}}\quad \textit{L}_p(\{\bf{W}_i\}, \{\bf{Y}_i^k\}, \{\bf{U}_i^k\})
\label{itera1}
\end{equation}
\begin{equation}
\small
    \bf{Z}_{i}^{k+1} := \underset{ {\bf{Z}}_{i}}{\text{arg min}}\ \textit{L}_p(\{\bf{W}_i^{k+1}\}, \{\bf{Y}_i\}, \{\bf{U}_i^k\})
\label{itera2}
\end{equation}
\begin{equation}
\small
    \bf{U}_{i}^{k+1} := \bf{U}_{i}^{k}+\bf{W}_{i}^{k+1}-\bf{Y}_{i}^{k+1}
\label{itera3}
\end{equation}
which (\ref{itera1}) is the proximal step, (\ref{itera2}) is projection step and (\ref{itera3}) is dual variables update. However, due to the non-convexity of the DNN loss function rather than the quadratic term in our method, the global optimality cannot be guaranteed.

\begin{figure}[t]
     \centering
     \includegraphics[width=0.38\textwidth]{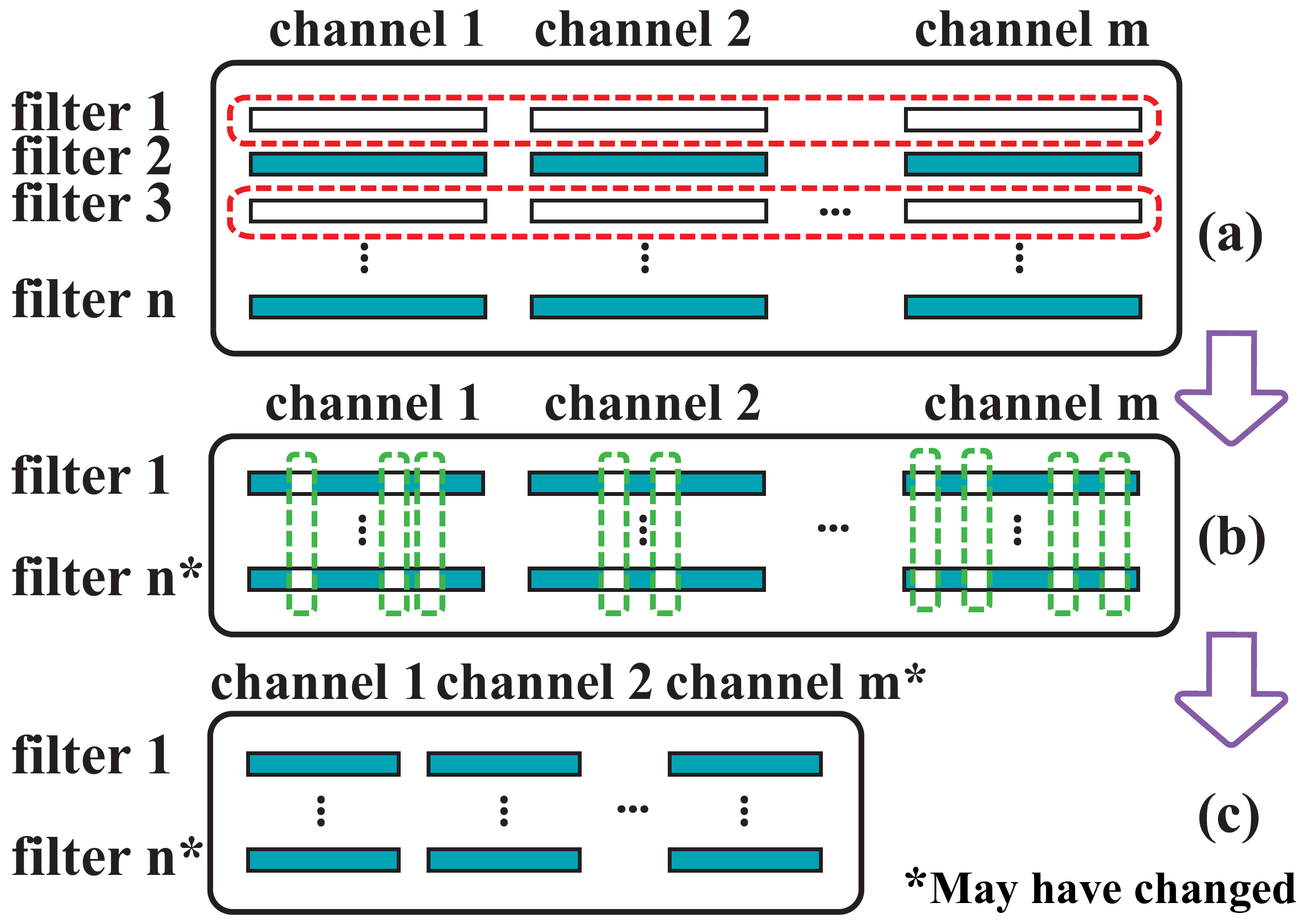}    
     \vspace{-4mm}
     \caption{GEMM view of weight pruning}
     \label{fig:gemm}
     \vspace{-3mm}
\end{figure}

Figure~\ref{fig:gemm} illustrate a combined structured pruning techniques in General Matrix Multiply (GEMM) view. We adopt filter pruning and column pruning together to reduce matrix dimension. As a result shows in Figure~\ref{fig:gemm} (c), the weight matrix size is reduced drastically compared with the original one, in the meantime, the shape of the weight matrix is still regular.

\begin{figure}[b]
     \centering
     \includegraphics[width=0.38\textwidth]{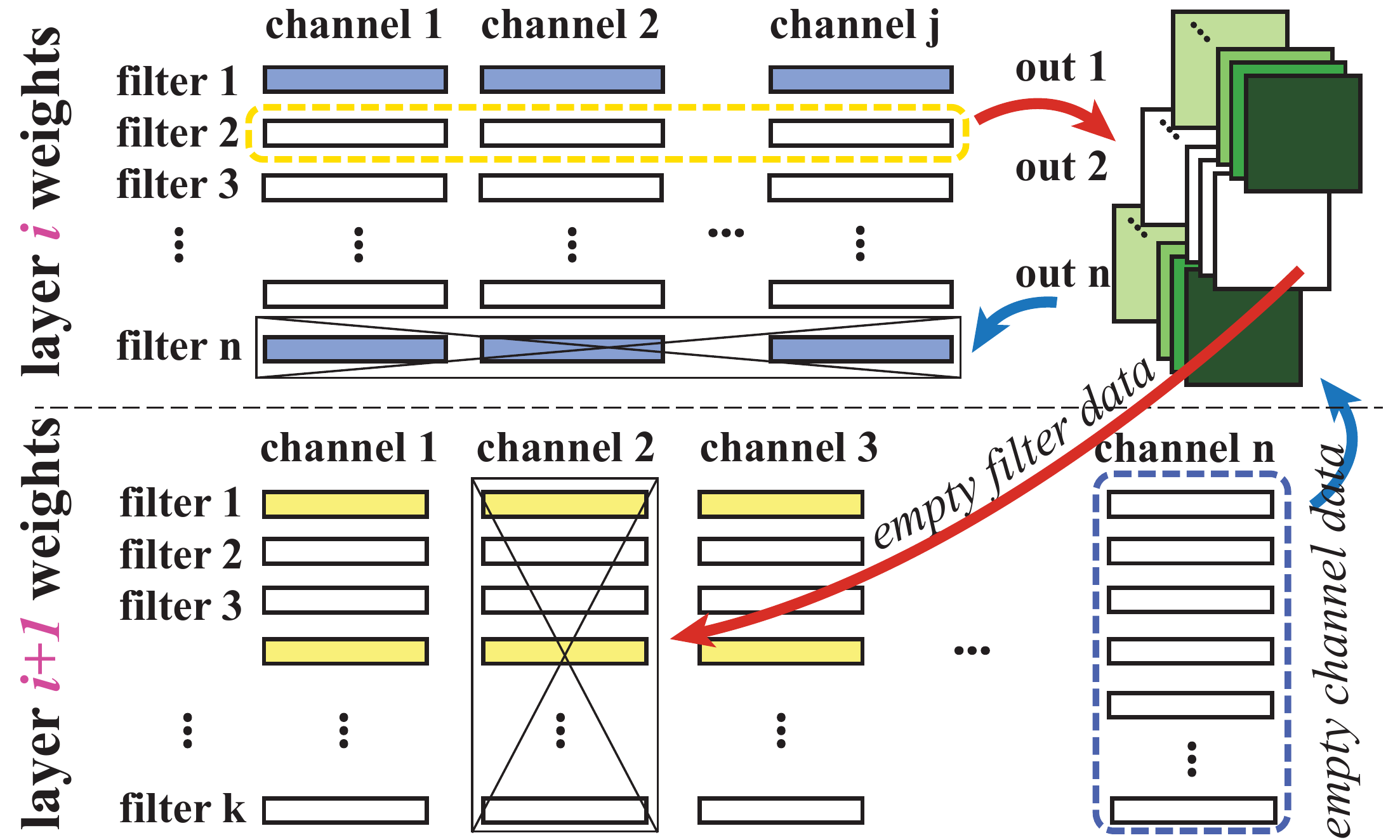}    
     \vspace{-3mm}
     \caption{Unused data path caused by structured pruning}
     \label{fig:unused}
     \vspace{-3mm}
\end{figure}

\section{Network Purification and Unused Path Removal (P-RM)}
ADMM weight pruning can significantly reduce weights while maintaining high accuracy. However, does the pruning process really remove all unnecessary weights?

From our observation and analysis on the data flow through a network, we find that if a whole filter is pruned, then after GEMM, the generated feature maps by this filter will be all ``blank". If those ``blank" feature maps input to next layer, then no matter what values are in the corresponding channel for those feature maps, the GEMM result will be zero. As a result, that channel will become unused channel which can be removed. By the same token, if a channel is pruned, then no matter what values are in the previous layer's corresponding filter, the GEMM result of the generated feature maps by this channel will be all zeros, in which case make that corresponding filter an unused one. Figure~\ref{fig:unused} gives a clear illustration about the corresponding relationship between the ADMM pruned filters/columns and the correspond unused channels/filters.

We further improved the empty channels caused unused filters method by creating a more generalized criterion that define what is ``emptiness" of a channel. Suppose $\Lambda_i$ is the number of columns per channel in layer $i$, and $\eta_{i,j}$ is the emptiness ratio. We have 
\vspace{-3.1mm}
\begin{equation}
    \eta_{i,j} = \big[\sum_{k=1}^{\delta} (column_k !=0)\big] / \delta \quad \delta \in \Lambda_i
    \label{equ:empty_ratio}
\vspace{-3.1mm}
\end{equation}
If $\eta_{i,j}$ exceed a pre-defined threshold, we can assume that this channel is empty. But this indiscriminate criterion has its limitation. The reason is that after pruning, the remaining columns are remained for a reason which is they are relatively ``important" to the whole network. If we remove all columns that satisfy $\eta$, disastrous accuracy drop will occur and hard to recover by retraining. 

In order to make our previous assumption work, we design a unified algorithm called ``\textit{Network Purification}" which is targeted on dealing with the \textit{non-optimality} problem of the ADMM process. By solving the problem, the above assumption can be validated simultaneously. We add a criterion constraint to compare the importance of the remaining columns channel-wisely and to help decide which columns can be sacrifice and which can not. 
We set-up an criterion constant $\sigma_{i,j}$ to represent channel $j$'s importance score, which is derived from an accumulation procedure:
\vspace{-3mm}
\begin{equation}
\vspace{-2.3mm}
    \sigma_{i,j} = \sum_{k=1}^{\delta}\ \|column_k\|_F^2 / \delta \quad \delta \in \Lambda_i
    \label{equ:importance_score}
\end{equation}
One can think of this process as if \textit{collection evidence for whether each channel that contains one or several columns need to be removed}. \textit{Network Purification} also works on purifying remaining filters and thus remove more unused path in the network. The effect of the combinatorially using \textit{Network Purification} and \textit{Unused Path Removal} (P-RM) is network will achieve extremely high compression rate without having any accuracy drop. Algorithm \ref{algo_unused} shows our generalized method of the P-RM method where $Th_1 \ldots Th_4$ are hyper-parameter thresholds values.
\begin{algorithm}[!h]
\footnotesize
\SetAlgoLined
\KwResult{Redundant weights and unused paths removed}
 Load ADMM pruned model\par
 $\delta$ = numbers of columns per channel\par
 \For{$i \gets 1$ until last layer}{
  \For{$j \gets 1$ until last $channel$ in $layer_i$}{
    \For{\normalfont{\textbf{each: }} $k \in \delta\ \normalfont{\textbf{and}}\ \|column_k\|_F^2 < Th_1$}{
        \normalfont{calculate: } $equation$ (\ref{equ:empty_ratio}), (\ref{equ:importance_score})\;
    }
    \If{$\eta_{i,j} < Th_2$\ \normalfont{\textbf{and}}\ $\sigma_{i,j} < Th_3$}{
        prune($channel_{i,j}$)\par
        prune($filter_{i-1,j}$) \ \bf{when} $i\neq1$\;
    }
    
  }
  \For{$m \gets 1$ until last $filter$ in $layer_i$}{
    \If{$filter_m$ \normalfont{\textbf{is}} empty \normalfont{\textbf{or}} $\|filter_m\|_F^2 < Th_4$}{
        prune($filter_{i,m}$)\par
        prune($channel_{i+1,m}$) \ \bf{when} $i\neq$ \normalfont{last layer index}\;
    }
  }
 }
 \caption{Network purification \& Unused path removal}
 \label{algo_unused}
\end{algorithm}

\vspace{-5mm}
\section{Experimental Results}
\begin{figure}[t]
     \centering
     \includegraphics[width=0.48\textwidth]{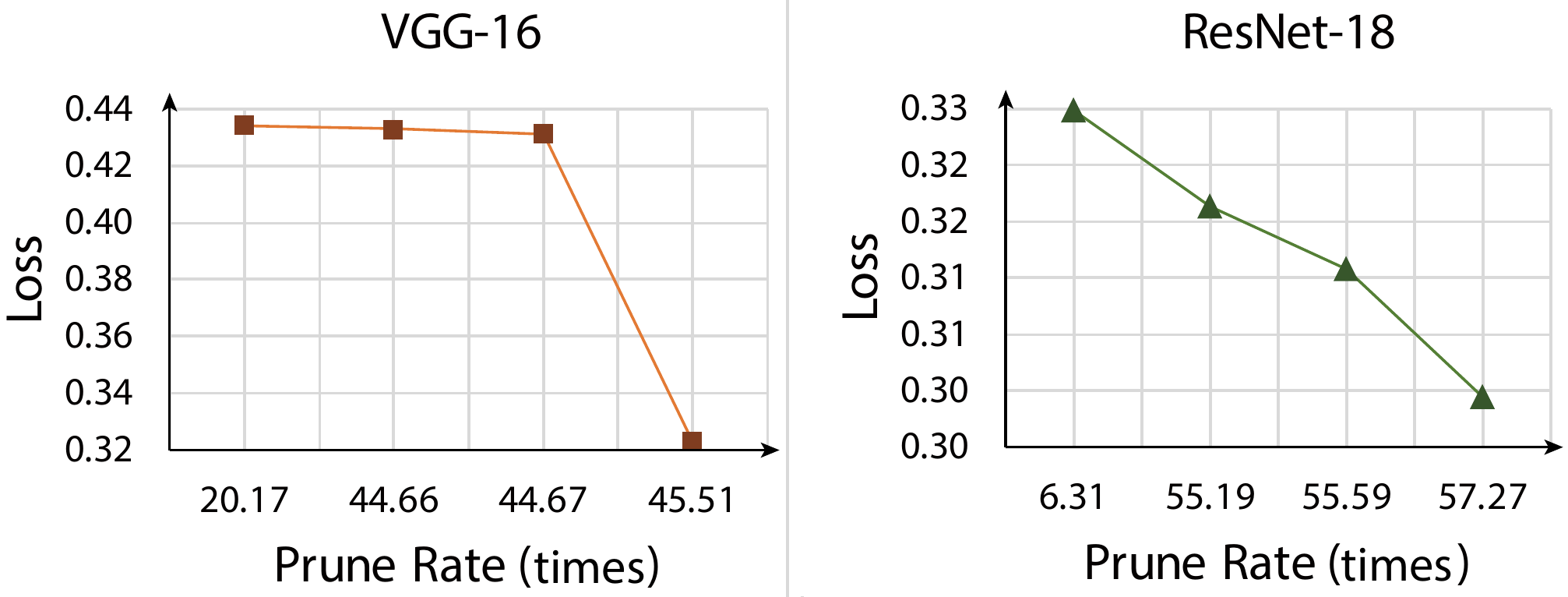}  
     \vspace{-6mm}
     \caption{Effect of removing redundant weights and unused paths. (dataset: CIFAR-10; Accuracy: VGG-16-93.36\%, ResNet-18-93.79\%. No retraining used)}
     \label{fig:purify_loss}
     \vspace{-5mm}
\end{figure}

Figure \ref{fig:purify_loss} proves that ADMM's non-optimality exists in a structured pruned model. By purifying the redundant weights, we can further optimize the loss function. All of the results are based on \emph{non-retraining} \textit{Network Purification} process. The purification along with removal of unused path (P-RM) process has great compression boost effect when the network is deep enough.

\begin{table}[tbh]
    \centering
    \vspace{-5mm}
    \caption{Structured weight pruning results on multi-layer network on MNIST, CIFAR-10 and ImageNet ILSVRC-2012 datasets}
    \renewcommand{\arraystretch}{1}
    \resizebox{0.5\textwidth}{!}{
        \begin{tabular}{c c c c c c} 
					\hline\hline
					\multicolumn{6}{c}{\textbf{Structured Weight Pruning Statistics}}  \\ 
					\hline
					\multirow{2}{*}{Method} & \multirow{2}{*}{\makecell{Original \\ Accuracy}} & \multirow{2}{*}{\makecell{Prune Rate \\ w/o P-RM}} & \multirow{2}{*}{\makecell{Accuracy \\ w/o P-RM}} & \multirow{2}{*}{\makecell{Prune Rate \\ with P-RM}} & \multirow{2}{*}{\makecell{Accuracy \\ with P-RM}}  \\ \\
					\hline
					\multicolumn{6}{c}{\textbf{MNIST}} \\
					\hline
					SSL &  & 26.10$\times$ & 99.00\% & N/A & N/A \\
					\hline
					\multirow{3}{*}{\makecell{\textbf{our} \\ \textbf{LeNet-5}}} & \multirow{3}{*}{99.17\%} & 23/18$\times$ & 99.20\% & \bf{39.23}$\times$  & \bf{99.20}\%  \\ 
					 &  & 34.46$\times$ & 99.06\% & \text{*}\bf{87.93}$\times$    & \bf{99.06}\%  \\ 
					 &  & 45.54$\times$ & 98.48\% & \bf{231.82}$\times$    & \bf{98.48}\%  \\ 
					\hline
					\multicolumn{6}{r}{\text{*}numbers of parameter reduced: \textbf{25.2K}}  \\
					\hline
					\multicolumn{6}{c}{\textbf{CIFAR-10}} \\
					\hline
					2PFPCE & 92.98\% & 4.00$\times$ & 92.76\% & N/A & N/A \\
					\hline
					\multirow{2}{*}{\makecell{\textbf{our} \\ \textbf{VGG-16}}} & \multirow{2}{*}{93.70\%} & \multirow{2}{*}{20.16$\times$} & \multirow{2}{*}{93.36\%} & \bf{44.67}$\times$ & \bf{93.36}\%  \\
					 &  &  &  & \text{*}\bf{50.02}$\times$ & \bf{92.73}\%  \\ 
					 \hline
					 AMC & 93.53\% & 1.70$\times$ & 93.55\% & N/A & N/A \\
					 \hline
					 \multirow{2}{*}{\makecell{\textbf{our} \\ \textbf{ResNet-18}}} & \multirow{2}{*}{94.14\%} & 5.83$\times$ & 93.79\% & \bf{52.07}$\times$  & \bf{93.79}\%  \\ 
					 &  & 15.14$\times$ & 93.20\% & \text{*}\bf{60.11}$\times$ & \bf{93.22}\%  \\ 
					 \hline
					 \multicolumn{6}{r}{\text{*}numbers of  parameter reduced on:}\\
					  \multicolumn{6}{r}{\textit{VGG-16}: \textbf{14.42M}, \textit{ResNet-18}: \textbf{10.97M}} \\
					 \hline
					\multicolumn{6}{c}{\textbf{ImageNet ILSVRC-2012}} \\
					\hline
					SSL AlexNet & 80.40\% & 1.40$\times$ & 80.40\% & N/A & N/A  \\
					\hline
					\bf{our AlexNet} & 82.40\% & 4.69$\times$ & 81.76\% & \bf{5.13}$\times$ & \bf{81.76}\% \\
					\hline
					\bf{our ResNet-18} & 89.07\% & 3.02$\times$ & 88.41\% & \bf{3.33}$\times$ & \bf{88.47\%}  \\
					\hline
					\bf{our ResNet-50} & 92.86\% & 2.00$\times$ & 92.26\% & \bf{2.70}$\times$ & \bf{92.27}\% \\
					\hline
					\multicolumn{6}{r}{numbers of  parameter reduced on:} \\
					\multicolumn{6}{r}{\textit{AlexNet}: \textbf{1.66M}, \textit{ResNet-18}: \textbf{7.81M}, \textit{ResNet-50}: \textbf{14.77M}} \\
					
					 \hline\hline
					 
        \end{tabular}
    }
    \label{table:results}
\end{table}

Table\ref{table:results} shows our experimental results of network pruning on Lenet-5, VGG-16, AlexNet and ResNet-18/50. The accuracy and pruning ratio results of SSL~\cite{wen2016learning} method is compared with our structured pruned Lenet-5 and AlexNet model, and 2PFPCE~\cite{min20182pfpce} and AMC~\cite{he2018amc} (ResNet-50) methods are compared with our VGG-16 and ResNet-18 on CIFAR-10 results. By using ADMM structured prune, \textit{Network Purification} and \textit{Unused Path Removal} (P-RM), LeNet-5 achieve 39$\times$ compression rate without accuracy drop, 88$\times$ compression with negligible accuracy drop and 232$\times$ with only 0.7\% accuracy drop. On CIFAR-10 dataset, our VGG-16 compressed model achieves 44$\times$ compression without accuracy degradation and 50$\times$ with 1\% accuracy drop and our ResNet-18 achieve 52$\times$ compression without noticeable accuracy loss and 60$\times$ compression with 0.9\% accuracy loss.

On ImageNet dataset, we increase AlexNet compression rate from 4.69$\times$ to 5.13$\times$, ResNet-18 from 3.02$\times$ to 3.33$\times$ and ResNet-50 from 2$\times$ to 2.7$\times$. All of our compression rate boost doesn't cause noticeable accuracy degradation.

\section{Conclusion}
In this paper, we provide an ADMM regularized method to achieve highly compressed DNN models with combination of different weight pruning structures, and maintain the network accuracy in a high level. We further investigate the post-process of ADMM pruning to solve the non-optimal solution caused by non-convex DNN loss function. We proposed \textit{Network Purification} and \textit{Unused Path Removal} that increase our model compression rate significantly.



\end{document}